# A Deep Learning-based Approach for Real-time Facemask Detection

Wadii Boulila
Robotics and Internet-of-Things Lab
Prince Sultan University
Riyadh, Saudi Arabia
National School of Computer Sciences
University of Manouba
Manouba, Tunisia

Ayyub Alzahem
College of Computer Science and Engineering
Taibah University
Medina, Saudi Arabia

Aseel Almoudi
College of Computer Science and Engineering
Taibah University
Medina, Saudi Arabia

Muhanad Afifi
College of Computer Science and Engineering
Taibah University
Medina, Saudi Arabia

Ibrahim Alturki
College of Computer Science and Engineering
Taibah University
Medina, Saudi Arabia

Maha Driss
Security Engineering Lab
Prince Sultan University
Ryiadh, Saudi Arabia
National School of Computer Sciences
University of Manouba
Manouba, Tunisia

*Abstract*— The COVID-19 pandemic is causing a global health crisis. Public spaces need to be safeguarded from the adverse effects of this pandemic. Wearing a facemask becomes one of the effective protection solutions adopted by many governments. Manual real-time monitoring of facemask wearing for a large group of people is becoming a difficult task. The goal of this paper is to use deep learning (DL), which has shown excellent results in many real-life applications, to ensure efficient real-time facemask detection. The proposed approach is based on two steps. An off-line step aiming to create a DL model that is able to detect and locate facemasks and whether they are appropriately worn. An online step that deploys the DL model at edge computing in order to detect masks in real-time. In this study, we propose to use MobileNetV2 to detect facemask in real-time. Several experiments are conducted and show good performances of the proposed approach (99% for training and testing accuracy). In addition, several comparisons with many state-of-the-art models namely ResNet50, DenseNet, and VGG16 show good performance of the MobileNetV2 in terms of training time and accuracy.

## I. Introduction

In the shadow of the COVID-19 pandemic, facemask wearing becomes mandatory in many public places throughout the world, a helpful solution that has proven to be useful in safeguarding these places and reducing the spread of this pandemic. Several rules are set to force wearing a facemask in public and work places, which represent hotspots for the spread of this infection. However, not every individual is aware or compliant, thus risking his or her life and the lives of others by not wearing a mask. Real-time monitoring of facemask wearing for a large group of people is becoming a difficult task. Manual monitoring is in general hard to enforce because of the manpower needed to efficiently protect public spaces and to ensure that individuals are wearing masks correctly. Aside from the cost problems and managerial effort, the biggest problem is the health factor because a certain set of employees will be in contact with hundreds of people daily, which poses a risk of them acting as infection points, therefore we aim to eliminate the human factor contact.

Recently, deep learning (DL) has been used in many domains and solved many complex problems, providing, therefore, significant results [1-4]. DL allows analyzing and interpreting massive volumes of data in a fast and accurate way [5-7]. Therefore, in this paper, we are proposing an approach that would help enforce the face-mask policy and monitor it with ease in real-time videos. The proposed system will help the authorities and commercial spaces monitor facemasks easily and efficiently.

The novelty of this paper with regard to exiting works is proposing an efficient and accurate approach for real-time videos. The proposed approach provides accurate detection of facemask wearing and whether it is worn in an appropriate way or not in real-time. To do this, a complete dataset is collected using public datasets and also our own one. In addition, the MobileNetV2 is used as a DL architecture for facemask detection. This approach has the advantage to be fast and suited to edge devices, and it provides excellent results for object detection. The proposed solution can be implemented in real-world surveillance cameras in public areas to check if people are following rules and wearing marks. The solution can be easily implemented with minimum resources.

The organization for the rest of the paper is as follows. Section 2 reviews previous related works. Section 3 illustrates the proposed approach for real-time face-mask detection. Section 4 reports and analyses the experimental results, and Section 6 presents the conclusions and future work.

## II. Related Works

This section presents the most recent and relevant academic works related to face-mask detection based on DL models. In [8], the authors propose to use the Faster Region-Based Convolutional Neural Network (R-CNN) algorithm to detect masks and to monitor social distancing. The proposed training model for mask detection is based on Single-Shot Multibox Detector (SSD) and You Only Look Once (YOLO) version 2. The testing of this model is performed on complex images including face turning, wearing classes, beard faces, and scarf images. The testing accuracy for this model attains 93.4%.

Convolutional Neural Network (CNN) model is also adopted by [9] to ensure real-time face mask detection. In this work, the authors propose to classify images provided in real-time by a camera connected to a Raspberry-Pi device. In this study, the authors developed a precise and speedy facemask detection. Training is conducted on a dataset composed of 25,000 images and an accuracy of 96% is achieved for training. The authors in [10] proposed a monitoring system allowing to detect if persons in a considered area are wearing or not a facemask and notify those who are not wearing a facemask by text messages. The proposed system uses Kaggle datasets for the detection model training and testing. The facemask is extracted from real-time faces of persons in public areas and is fed as an input into a CNN. The real-time automated facemask detection is performed using MobileNet and OpenCV. Nagrath et al. [11] suggested a model named SSDMNV2 for face mask detection using OpenCV Deep Neural Network (DNN), TensorFlow, Keras, and MobileNetV2 architecture. SSDMNV2 allows classifying whether a person in an image is wearing a facemask or not. It uses a pre-trained deep learning model called MobileNetV2, which is a variant of CNN. The classification phase is preceded by a face detection phase that is performed by DNN. The average accuracy of the proposed model is 93 %. Transfer learning of InceptionV3 is used in [12] to detect persons who are not wearing a facemask in public places by integrating it with surveillance cameras. To enhance the performance of the proposed model by increasing the diversity of the training data, image augmentation techniques are used. The training images are augmented by applying eight different operations, which are: shearing, contrasting, flipping horizontally, rotating, zooming, and blurring. The proposed transfer-learning model achieved an accuracy of 99.92% during training and 100% during testing on the considered dataset. In [13], a new model based on YOLO v2, ResNet-50, and Adam optimizer is used. To train and validate the proposed detection model, a new dataset based on two public masked face datasets is considered. The experiments that are conducted on this new dataset have shown that the proposed model provides an average precision equal to 81%.

Most existing facemask detection approaches focus on whether the person is wearing a facemask or not. In our work, we propose to detect facemask wearing in real-time videos. In addition, the proposed system achieves excellent accuracy compared to existing works and it allows detecting whether a facemask is appropriately worn or not.

### III. PROPOSED APPROACH

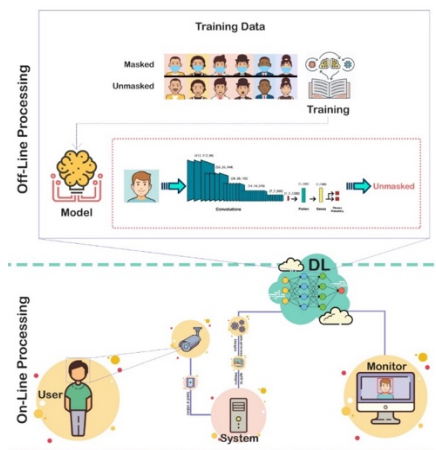

Fig. 1. Proposed Approach

The proposed approach is based on two phases: online and offline processing. The purpose of the off-line processing is to create a DL model that is able to detect and locate facemasks and whether they are appropriately worn. The online phase aims to deploy the DL model on edge computing in order to detect masks in real-time. Figure 1 depicts the two phases of the proposed approach.

*A. CNN architecture*

In this paper, we propose to use MobileNetV2 architecture to ensure accurate face-mask detection, the proposed CNN architecture is presented in Figure 2. We choose to use MobileNetV2 since it provides several advantages such as 1) it is a light-weight DL suited to edge devices, 2) it provides excellent results for object detection, and 3) it can efficiently tradeoff between accuracy and latency using simple global hyperparameters [14].

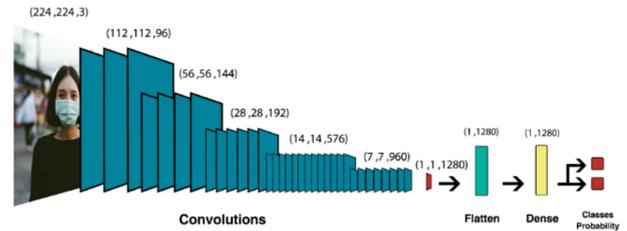

Fig. 2. CNN architecture

*B. Data Augmentation*

In order to have more diversity in our dataset, we used data augmentation. Several transformations are modes such as geometric transformations, flipping, color modification, width weight compensation, and rotation.

Figure 3 shows some transformations made to our dataset.

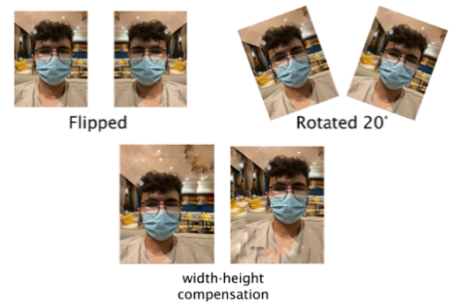

Fig. 3. Data Augmentation

### IV. EXPERIMENTS

In this section, we will start by presenting the dataset used in this study. Then, many experiments are conducted to show the good performances of the considered CNN model in detecting facemask in real-time videos. Finally, several comparisons with many state-of-the-art models show good performance of the considered Mo-bileNetV2 in terms of training time and accuracy.

*A. Dataset Description and Implementation Details*

In order to build a complete and diversified dataset related to facemask, we combined public datasets with our own dataset. The public datasets used in this study are: 1) Face Mask Lite Dataset [1], Real-World-Masked-Face-Dataset [2],

---

[1] https://www.kaggle.com/prasoonkottarathil/face-mask-lite-dataset

[2] https://github.com/X-zhangyang/Real-World-Masked-Face-Dataset

MaskedFace-Net[3], and face mask detection[4]. The obtained dataset contains both people wearing masks, without masks, wearing masks in an inappropriate way. The final dataset has an initial size of 60 GB and contains 21407 images. Table 2 describes the number of images with/without a mask for training and testing.

To train the our model, images were converted individually into an array and then preprocessed by using the "preprocess_input" in "tensorflow.keras.mobilenet". after that, images were stored into a list, after preprocessing all images, the list was converted into a NumPy array as a floot32 datatype.

TABLE I. NUMBER OF IMAGES FOR TRAINING AND TESTING

| Number of images | Training | Testing | Total |
|---|---|---|---|
| With mask | 8079 | 2020 | 10099 |
| Without mask | 9046 | 2262 | 11308 |
| **Total** | 17125 | 4282 | **21407** |

In this study, the experiments are carried out using a PC with the following configuration properties: an x64-based processor, an Intel Core i7-9750 CPU @2.60GHz, and a 16 GB RAM running on Windows 10 with NVIDIA GeForce GTX 1650. The CNN architectures are programmed using Jupyter notebook under python 3.7 programming language. We used both the Keras library and TensorFlow backend. For faster computation, we used Collaboratory pro with 25.51 GB RAM and 225.89 GB Disk, connected to python 3 Google Compute Engine backend.

*B. Results*

We trained the proposed model with 32 batch sizes and 100 epochs and it took almost 12 hours to complete the training. After training the model with the first dataset version, we test it to see the actual results. Most of the wrong results across because of conflict images, not enough images for a specific angle of the face sides, or sometimes there is not an image for a specific angle. Therefore, after every dataset version, we return to evaluate results and manually remove/add images with the required angles to cover all the face sides and angles. We iteratively repeat this process until reaching satisfying results and that happened after the eighth version of the dataset.

To minimize the false positive and negative rate in the execution as possible, we add a test that ensures that four consecutive frames from the video have the same state (either the person is wearing a mask or not). This will reduce the error of detecting masks from a different angle of view.

```
if (mask < withoutMask):
    fn += 1
    if fn > 2 or fn == -1:
        label = "No Mask"
        color = (0, 0, 255)
        tn = 0
    if fn == 4:
        winsound.Beep(freq, duration) # alert
else:
    tn += 1
    if tn > 2 or tn == -1:
        label = "Mask"
        color = (0, 255, 0)
        fn = 0
```

Fig. 4. Code for removing errors related to moving people

The code that we added is depicted in Figure 4. This code helped to improve the accuracy of detecting a person either is wearing a mask or not even from different angles of view of this person.

To evaluate the performances of the proposed CNN architecture, several performance metrics are used namely the accuracy, precision, recall, and F1-score. These measures are computed using four variables: true positive (TP), true negative (TN), false positive (FP), and false-negative (FN). TP variable denotes an outcome where the model correctly predicts the positive class. TN variable denotes an outcome where the model correctly predicts the negative class. FP denotes an outcome where the model incorrectly predicts the positive class. Finally, FN denotes an outcome where the model incorrectly predicts the negative class. Accuracy, precision, recall, and F1-score are calculated using the equation 1, 2, 3, and 4 respectively.

$$Precision = \frac{tp}{tp+fp} \quad (1)$$

$$Recall = \frac{tp}{tp+fn} \quad (2)$$

$$Accuracy = \frac{tp+tn}{tp+tn+fp+fn} \quad (3)$$

$$F1-score = 2*\frac{precision*recall}{precision+recall} \quad (4)$$

In Figure 5, the classification report of the proposed CNN architecture is depicted. We can conclude that our model achieves excellent values in detecting facemask with 99% accuracy, precision, recall, and F1-score.

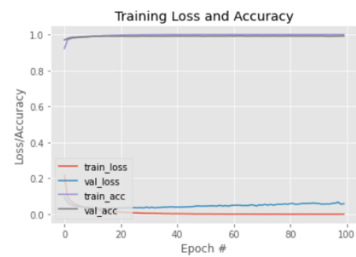

Fig. 5. Classification Report

To evaluate the performance of the proposed CNN model, the training and validation loss and accuracy functions are computed. The model was trained with 32 batch sizes, 100 epochs, and the training time is almost 12 hours. Figure 6 describes the training and validation accuracy and loss of our model.

Fig. 6. Training/Validation accuracy and loss

| 0.99 | 0.01 |
|---|---|
| 0.01 | 0.99 |

Fig. 7. Confusion Matrix

In addition, in order to describe the classification performance of the CNN model, the confusion matrix is plotted in Figure 7. We have two classes that are with-mask and without-mask.

---

[3] https://github.com/cabani/MaskedFace-Net

[4] https://www.kaggle.com/andrewmvd/face-mask-detection

## C. Evaluation of the proposed approach

The considered MobileNetV2 CNN model is compared to state-of-the-art DL models namely DenseNet, ResNet50, and VGG16. They are trained with 32 and 20 epochs to compare their validation accuracy and loss on the same dataset. The training time is depicted in Table 3.

COMPARISON OF THE TRAINING TIME OF DENSENET, RESNET50, VGG16, AND MOBILENETV2

| DL model | Training time (Hours: minutes) |
|---|---|
| DenseNet | 4:24 |
| ResNet | 5:58 |
| VGG | 18:51 |
| **MobileNetV2** | **1:26** |

```
              precision    recall  f1-score
   with_mask       0.93      0.71      0.80
without_mask       0.94      0.99      0.96

    accuracy                           0.94
   macro avg       0.93      0.85      0.88
weighted avg       0.94      0.94      0.94
```
ResNet50

```
              precision    recall  f1-score
   with_mask       0.97      0.90      0.94
without_mask       0.99      1.00      0.99

    accuracy                           0.98
   macro avg       0.98      0.95      0.96
weighted avg       0.98      0.98      0.98
```
VGG16

```
              precision    recall  f1-score
   with_mask       0.98      0.97      0.98
without_mask       1.00      1.00      1.00

    accuracy                           0.99
   macro avg       0.99      0.98      0.99
weighted avg       0.99      0.99      0.99
```
DenseNet

```
              precision    recall  f1-score
   with_mask       0.98      0.98      0.98
without_mask       1.00      1.00      1.00

    accuracy                           0.99
   macro avg       0.99      0.99      0.99
weighted avg       0.99      0.99      0.99
```
MobileNetV2

Fig. 8. Comparison of performance metrics between ResNet50, VGG16, DenseNet, and MobileNetV2

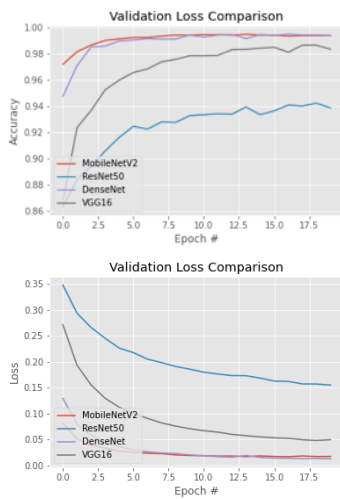

Fig. 9. Comparison of validation accuracy and loss between MobileNetV2, ResNet50, DenseNet, and VGG16

Finally, a comparison of the validation accuracy and loss results between MobileNetV2, ResNet50, DenseNet, and VGG16 is depicted in Figure 15. Results show that the MobileNetV2 model has the best results, which justifies the use of this model in our case.

## V. CONCLUSION

In this paper, an efficient CNN model based on MobileNetV2 for Real-time Facemask Detection is presented. The proposed approach achieved 99% accuracy in training and testing and can determine whether a mask is appropriately worn or not in real-time video streams. Extensive experiments are conducted to show the good performances of the MobileNetV2 model in detecting facemask in real-time videos. In addition, several comparisons with many state-of-the-art models namely ResNet50, DenseNet, and VGG16 show good performance of the MobileNetV2 in terms of training time and accuracy.

As future work, further experiments will be conducted to evaluate the performances of the proposed solution. In addition, we plan to implement the proposed solution in real-world surveillance cameras in public areas to check if people are following rules and wearing masks appropriately.